8

# AI-based Classification of Customer Support Tickets: State of the Art and Implementation with AutoML

Mario Truss and Stephan Böhm

*Center of Advanced E-Business Studies*
*RheinMain University of Applied Sciences, Wiesbaden, Germany*
*e-mail: itsmariotruss@gmail.com; stephan.boehm@hs-rm.de*

*Abstract*—Automation of support ticket classification is crucial to improve customer support performance and shortening resolution time for customer inquiries. This research aims to test the applicability of automated machine learning (AutoML) as a technology to train a machine learning model (ML model) that can classify support tickets. The model evaluation conducted in this research shows that AutoML can be used to train ML models with good classification performance. Moreover, this paper fills a research gap by providing new insights into developing AI solutions without a dedicated professional by utilizing AutoML, which makes this technology more accessible for companies without specialized AI departments and staff.

# 1. Introduction

One of today's primary priorities of companies is to improve the Customer Experience (CX) to increase customer satisfaction and reduce churn. However, "just 2 percent of organizations reached the top stage of CX maturity [and] most organizations are in early stages of CX maturity" (Dorsey et al., 2022). According to a recent study by Qualtrics (2022), 47 percent of customers ranked support as the second most important area of improvement in CX. One major factor of customer satisfaction identified in recent research (e.g., Service Excellence Research Group, 2021) is the speed at which customer support answers customer inquiries. Demand for customer support is rising and often exceeds the supply of available support agents. Especially missing knowledge and multiple re-routings between support agents are major factors for delays in resolution time. Further research suggests that due to information overload, the quality of decisions decreases with the number of decisions (Hemp, 2009; Viegas et al., 2015). In most recent studies, lack of time and resources are mentioned as the main issues in customer support, which harm the performance and, ultimately, the customer experience (HubSpot, 2022; Serrano et al., 2021). All of this underlines the need for customer support automation. In this context, this research aims to analyze whether automated machine learning (AutoML) can be used to automate the support ticket classification process. Given the novelty of the topic, we attempt to answer the following research questions with our study:

RQ1: *What is the state of research on AutoML-based ticket classification?*

RQ2: *Can standard AutoML solutions be applied to support ticket classification with usable results?*

RQ3: *What are the requirements for the necessary training data?*

RQ4: *What implications can be derived for ML model development?*

To answer these research questions, this paper is structured as follows: The first section describes customer support classification problems, specifically ticket classification. Additionally, relevant technological concepts of artificial intelligence, machine learning, and automated machine learning are discussed. The second section summarizes the state-of-the-art of AI-based approaches to address these problems. Section three gives insights into the methodological approach of the AutoML results assessment by defining evaluation criteria. Finally, the last two sections present the findings, conclusions, implications, and limitations.

## 2.    Research Background

### 2.1    Support Ticket Classification Problems

A common practice in customer support is using a support ticket system (STS), where customers can create support tickets. In the tickets, customers describe and document the issue they face, which will then be read and classified by customer support, e.g., assigning the ticket to a category, priority, or other attributes. The correct classification is needed for a quick solution to the issue by the right person. Figure 8.1 shows an exemplary classification process for a support ticket. A support agent usually does ticket classification manually to allow for resource allocation (Bruton, 2002).

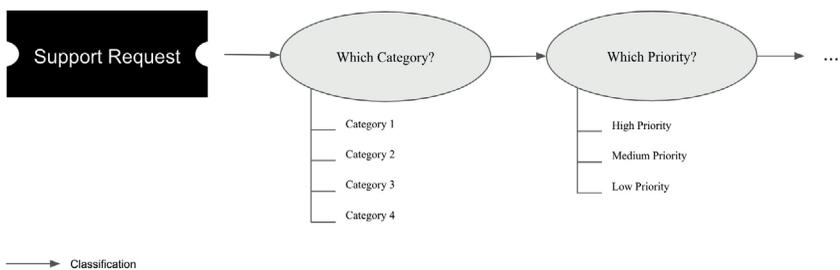

Figure 8.1: Support Ticket Classification Process

This manual classification process directly bottlenecks the resolution time of the support request and, therefore, can negatively influence the overall customer experience through increased customer waiting time (Service Excellence Research Group, 2021). As a result, companies invest in automating this classification process to reduce waiting time and workload on support agents (Service Excellence Research Group 2021; McKinsey, 2022a).

The management of support tickets in Information Technology (IT) is often referred to as IT Service Management (ITSM). However, commercial solutions for AI-based ticket classification are not widely available for all STS yet. So far, only a few AITSM solutions (e.g., ServiceNow, 2023) are available on the market. This study is based on the solution Jira by Atlassian (2023), which is widespread in IT support. When this study was conducted, no AI-based automatic classification was available in this solution. Thus, an external solution must be found for automating the ticket classification task.

## 2.2 Artificial Intelligence and Machine Learning

The term Artificial Intelligence (AI) stems from Alan Turing and John McCarthy and describes a machine with artificially created intelligence that enables the machine to autonomously perform cognitive tasks, usually associated with humans (Kirste & Schürholz, 2019; Klüver & Klüver, 2021; Turing, 1950). Machine learning is a form of AI that enables the analysis of data and the recognition of patterns based on algorithmic learning processes. The type of ML used for this paper is supervised learning, more specifically classification. In classification, patterns are algorithmically derived from training data, with input features and the resulting output classes, to learn the prediction of classifications for new data. The trained intelligence for this purpose is stored in an ML model (Figure 8.2) and is derived through ML algorithms. In contrast to ML, deep learning (DL) is an advanced form of ML that utilizes neural networks (Buxmann & Schmidt, 2019; Kirste, 2019; Mondal, 2020; Thrun & Lorien, 1998). Due to the complexity and the not immediately comprehensible form of information processing, ML models based on neural networks are often called a 'black box' (Browne, 2022).

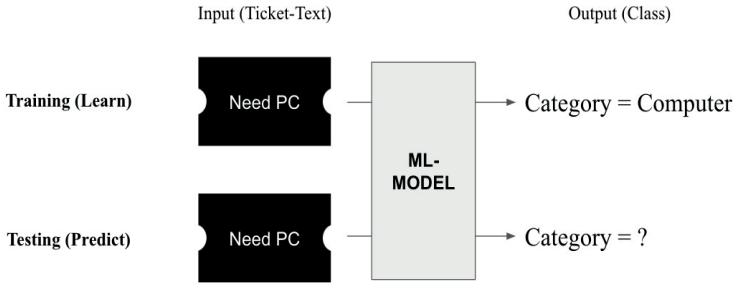

Figure 8.2: ML Model

In general, developing an ML model requires professionals with extensive technical knowledge, which can cause a deceleration in the ML model development. Due to the increasing labor shortage and a lack of qualified professionals in the job market, hiring AI specialists to develop ML models is a major challenge for companies (McKinsey, 2022b). This is especially true for small and medium-sized enterprises (SMEs) (Heizmann et al., 2022). Additionally, IBM (2022) research shows that the development of AI solutions continues to be mainly in the hands of IT specialists, hindering AI democratization and an interdisciplinary approach to development. Due to these circumstances, the advancement of the maturity of AI in companies is delayed and bottlenecked (McKinsey, 2022b).

## 2.3    Automated Machine Learning (AutoML)

Automated Machine Learning (AutoML) is an AI technology with which the development of ML models is to be simplified and automated as far as possible. AutoML was invented to democratize ML model development and enable non-technical professionals to develop or at least participate in developing AI solutions. AutoML automates manual ML model development steps, such as preprocessing, featureextraction, model selection, optimization, and evaluation (Nagarajah & Poravi, 2019). However, comprehensive support for ML model development is required, as the manual process is prone to errors and "entails long and careful design and tuning activities" (Zicari et al., 2022) to maximize performance. Therefore, a sophisti-cated user interface is offered to guide non-experts through the whole ML model development process. For example, this interface provides guidelines on what data to use and detailed requirements such as dataset size (Chen et al., 2019).

## 3.   Related Research

AI-based or, more specifically, ML-based ticket classification is not an entirely new research area. For example, several recent studies are looking at approaches to using ML to improve customer support (Fuchs et al., 2022b; Montgomery et al., 2018; Ulges et al., 2020) or reduce ticket classification error rates (Khowongprasoed & Titijaroonroj, 2022). Early research on applying ML-based classification in the context of IT support was performed by Jonsson (2013), Zhou et al. (2015), and Zinner et al. (2015).

Due to the fast pace of the developments in the fields of AI and ML, this paper only takes current research into account that was released after 2017 to ensure timeliness and comparability of results. With this restraint, a comprehensive literature search (query: AI *OR* classification *OR* ML *AND* customer support *OR* support ticket *OR* ITSM) was conducted in relevant literature databases (e.g., ACM, DL, IEEE, etc). As a result, six different types of classification were identified in the current research: Identify spam, assign, categorize, escalate, and prioritize tickets or to determine the sentiment of ticket texts. Most research focused on custom implementations with ML (e.g., Support Vector Machine, SVM, Random Forest, RF, Decision Tree, DF, XGBoost) or DL (e.g., Convolutional Neural Networks, CNN). Data scientists or ML engineers implemented almost all solutions with traditional coding practices. Additionally, rule engines were used as a complement to the ML model (Mandal et al., 2019).

In contrast to the more widespread traditional ML approaches, only a few papers have been published specifically on applying AutoML to support tickets. AutoML is only discussed twice in research for this purpose. For example, Yayah et al. (2021) discussed H20 as a solution to classify a "resolution code" for network issue tickets. Another paper by Jyotheeswar et al. (2020) discussed Azure ML as a solution for ticket classification based on a larger dataset of more than 46,000 tickets. The literature review shows that there is still a need for further research. Also, the authors are not aware of any research that explicitly uses the AutoML solution VertexAI by Google, which is the basis of our study, and applies it to a smaller dataset of tickets. However, the quality of ML models depends not only on the development process. The training data used also has a major influence on the quality of the model. Against this background, data-related problem areas must also be adequately considered when evaluating AutoML procedures. For this reason, problems in datasets that are relevant for ticket classification will be discussed below.

Research by Wahba et al. (2020) discussed class imbalance as a challenge for ML training. A study by Yang and Xu (2020) focused on the problem of "heavy class imbalance". This problem can occur in datasets with an overrepresentation of majority classes and an underrepresentation of minority classes. One example for a class imbalance would be, that in a classification between two classes, there are 5 example tickets for class "A", but 100 examples for class "B" in the training data. This can potentially cause that the ML model is optimized for classifying class "B", which is called a "majority class". Especially heavy class imbalance can "significantly affect the performance of machine learning models" and lead to underperformance "in minority classes", which is the class with a low amount of examples in the training data (Singh & Vanschoren, 2022). Researchers suggest manually reducing the majority classes or using oversampling techniques to address these issues, which would result in a balanced dataset without class imbalance. Oversampling describes synthetically creating new examples within minority classes (Wahba et al., 2020). According to supporting research, this can improve class imbalance, bias, and ML model performance (Nisbet et al., 2017).

Another data-related problem area seems to be the ambiguity of the classes in the training data, which can lead to misclassification in practice. A notable approach to improving the uniqueness of the classes is to derive new classes with a technique called topic modeling (Goby et al., 2016).

It is worth mentioning that most research we identified in our review uses larger datasets for classification model training (average: 101,578; median: 16,364). This fact might create the impression that large datasets lead to better results. Contrarily, research shows using reduced, but selective datasets can lead to better performance (Revina et al., 2020). Furthermore, according to supporting research, the quality of the classes and the balance have more impact on model performance than the sheer training set size (Wahba et al., 2020).

## 4. Methodology

### 4.1 Dataset and Preprocessing

The following analysis was based on IT customer support tickets of a German software company. The dataset was provided by the IT customer support unit of the company and exported from the STS Jira. The data structure of the tickets included nine data fields, as shown in Table 8.1 below. For further processing with Python, Pandas, NLTK, Jupyter Notebook and finally AutoML, the tickets were exported as a CSV file.

Table 8.1: Support Ticket Field

| Field | Data Type | Example |
|---|---|---|
| Creator | Text | John Doe |
| Support Agent | Text | Mario Truss |
| Creation Date | Date | 2022-01-01 |
| Title | Text | PC is broken |
| Description | Text | My pc is broken |
| Attachment | File, e.g., jpg, pdf | screenshot.jpg |
| Category | Text | Repair |
| Priority | Text | High |
| Escalation | Text | Escalated / Don't escalate |

Based on the relevance for the company's IT support team and previous research, two classification types have been selected for this study: Category (what category is it?) and escalation (will it escalate?). In addition, the evaluation of previous research has shown that the features "Title" and "Description" are essential for successful classification. For this reason, only these two data fields of the tickets from Table 8.1 were selected as input features for the model training.

An imbalanced and balanced dataset was created for each classification type (Table 8.2). No oversampling was applied, but instead, the majority classes were reduced to increase balance, as suggested in related research (Quiñonero-Candela et al., 2022). For the classification by category, the number of classes was reduced from 288 to the ten most used ("Inventory", "Offboarding", "Phone", "Training", "Admin", "Plugin", "Onboarding", "Purchase" and "Repair") to meet the requirements of the AutoML solution and reduce class ambiguity (Google, 2023). The resulting ML models trained will be referred to as M1, M2, M3, and M4, as shown in Table 8.2.

Table 8.2: Category ML Model Performance Metrics

| Model | Class Distribution | Classification | No. of Tickets |
|-------|-------------------|----------------|----------------|
| M1 | Imbalanced | Category | 1,573 |
| M2 | Balanced | Category | 996 |
| M3 | Imbalanced | Escalation | 4,265 |
| M4 | Balanced | Escalation | 3,443 |

To eliminate unwanted data from the tickets and to maintain comparability with traditional ML techniques (Wahba et al., 2020; Opuchlich, 2020), the following preprocessing steps with Trifacta Dataprep, Pandas, and NLKT were performed:

Concatenation of "Title" and "Description"

Elimination of stop words for removing irrelevant and privacy-related content (URL, emails, first and last name, special characters, IP address)

Lowercasing of all text

With these steps, the preparation of the data sets was completed. In the following section, we describe the development of the AutoML model based on the four datasets and the model evaluation.

## 4.2    Model Training and Evaluation Criteria

The AutoML solution used for this research was Google Vertex AI, as shown in Figure 8.3. The following process was repeated each of the models M1, M2, M3, and M4. First the dataset was selected, then the annotation set was automatically created, the objective was set to "Text classification (Single-label)" and the deployment method was "AutoML". Each dataset was randomly split (data split method) into 80 percent training data, 10 percent validation data for model tuning while training, and 10 percent test data for final testing.

Figure 8.3: Google VertexAI Configuration

VertexAI computed the three evaluation metrics Precision, Recall, and F1 Score for each ML model trained based on the four datasets. The meaning of these metrics can be explained in a simplified way using the formulas for an example with two classes (Positive and Negative):

$$Precision\ (P) = \frac{TP\ (True\ Positives)}{TP+FP\ (False\ Positives)} \qquad (1)$$

Precision measures the proportion of correctly classified instances compared to all instances predicted for a class.

$$Recall\ (R) = \frac{TP}{TP+FN\ (False\ Negatives)} \qquad (2)$$

Recall measures the proportion of correctly classified instances compared to all instances of the actual class.

$$F1\ Score\ (F1) = 2 * \frac{P*R}{P+R} \qquad (3)$$

F1 combines both metrics in a single value and represents the harmonic mean of P and R. Accuracy was not used as it is inappropriate for imbalanced data (Yang, X.-S., 2022). Auto ML models can be developed for high P, R, or F1, depending on the optimization objectives. Since no specific optimization goal was specified, the F1 score was chosen for model evaluation to enable comparability based on a combined metric. Inspired by Allwright (2022), the following grading schema was used to score the performance metrics, because there is no standard in research (Table 8.3).

Table 8.3: Category ML Model Performance Metrics

| Metric Score (P, R, F1) | 1 (100%) | > 0.9 (90%) | 0.8 - 0.9 (80 - 90%) | 0.5 - 0.8 (50-80%) | <0.5 (50%) |
|---|---|---|---|---|---|
| Grade | Perfect | Very good | Good | OK | Not good |

Additionally, a Confusion Matrix was utilized to quantify the incorrect prediction of classes compared to the correct class, to visualize the confusion with another class, as seen in Figure 8.4. Finally, to understand the root causes of misclassifications, incorrectly classified tickets were analyzed qualitatively.

| True Label | Predicted Label | | |
|---|---|---|---|
| | A | B | C |
| A | 80% | 10% | 10% |
| B | 0% | 100% | 0% |
| C | 10% | 0% | 90% |

Figure 8.4: Confusion Matrix

## 5. Findings and Performance Evaluation

### 5.1 Classification by Ticket Category

The research findings for the classification by ticket category in Table 8.4 show that balanced and imbalanced datasets led to a similar AutoML model performance. The M1 with an imbalanced dataset (1,573 tickets) achieved P = 0.877, R = 0.871, F1 = 0.844, and thus a good performance. For M2 with a balanced dataset (996 tickets), the results were slightly better: P = 0.891, R = 0.882, and F1 = 0.886. These metrics reveal that M2 with a balanced dataset resulted in higher classification performance (P → +1.4%, R → +1.1%, F1 → +1.2%), compared to M1 with the imbalanced dataset, despite using a smaller dataset.

Table 8.4: Category ML Model Performance Metrics

| Model | Class Distribution | Classification | No. of Tickets | P | R | F1 |
|---|---|---|---|---|---|---|
| M1 | Imbalanced | Category | 1,573 | 0.877 | 0.871 | 0.874 |
| M2 | Balanced | Category | 996 | 0.891 | 0.882 | 0.886 |

The confusion matrix (Figure 8.5) further illustrates that the performance of the ML model varied by class.

| M1 | Predicted Label | | | | | | | | |
|---|---|---|---|---|---|---|---|---|---|
| **True Label** | Inventory | Offboarding | Phone | Training | Admin | Plugin | Onboarding | Purchase | Repair |
| Inventory | 82% | | | | | | | 18% | |
| Offboarding | | 88% | | | 13% | | | | |
| Phone | | | 63% | | | | | 38% | |
| Training | | | | 100% | | | | | |
| Admin | | | | | 60% | | | 40% | |
| Plugin | | | | | | 100% | | | |
| Onboarding | | 10% | | | | 10% | 80% | | |
| Purchase | | | | | | | 4% | 91% | |
| Repair | | | | | | | | 43% | 57% |

Figure 8.5: Confusion Matrix Category (M1)

For M1, except for the classes "Training" and "Plugin", there were errors in all classifications (Figure 8.5). The most significant misclassification occurred for the following classes:

"Repair" (43 percent incorrect), confused with "Purchase".

"Admin" (40 percent incorrect), confused with "Purchase".

"Phone" (37 percent incorrect), confused with "Purchase".

For M2, the overall classifications were more frequently correct (Figure 8.6).

| M2 | Predicted Label | | | | | | | | |
|---|---|---|---|---|---|---|---|---|---|
| **True Label** | Inventory | Offboarding | Phone | Training | Admin | Plugin | Onboarding | Purchase | Repair |
| Inventory | 100% | | | | | | | | |
| Offboarding | | 100% | | | | | | | |
| Phone | | | 90% | | | | | | |
| Training | | | | 100% | | | | | |
| Admin | | | 20% | | 20% | | 20% | 20% | 20% |
| Plugin | | | | | | 100% | | | |
| Onboarding | | | | | | | 100% | | |
| Purchase | | | | | 15% | | | 85% | |
| Repair | | | 14% | | | | | | 57% |

Figure 8.6: Confusion Matrix Category (M2)

Still, the following classes got misclassified:

"Admin" (80 percent incorrect), confused with "Phone", "Onboarding", "Purchase", and "Repair".

"Repair" (43 percent incorrect), confused with "Phone".

"Purchase" (15 percent incorrect), confused with "Admin".

"Phone" (10 percent incorrect), confused with "Admin" and "Repair".

Further qualitative analysis of the tickets revealed that most misclassification cases occurred for classes where the training dataset of M1 and M2 contained ambiguous classes. This means, that a human used various classes for the same topic. Therefore, the classes used are not mutually exclusive. As ML models learn form examples, meaning classification made by humans, the ambigious classification will be learned and then be replicated in the ML-based classifications. For example, the following cases are observable, as seen in Table 8.5:

1. Tickets about a "mobile" were classified by humans both as "Phone" and "Inventory", but the ML model predicts "Phone" for both, because the content ("mobile") is too similar.

2. The need for a new device ("mobile" and "mouse") were classified both as "Purchase" and "Phone". As "Purchase" would also be a logical class for the request of "new mouse", there is an ambiguity in the training data.

3. The class "Admin" describes both cases for "access" to software, as well as problems with "VPN". Therefore, this class seems to be a very generic class.

4. The "need" for "access" is classified both as "Admin" and "Onboarding", which means that giving "access" is both an administrative, as well as an onboarding task.

5. The predicted class for the need to access a software is predicted by the ML model as both "Purchase" and "Admin". This means that in the training data, there were also cases where the "need access" request was also classified as a purchase. This shows that the classes revolving around access rights for software are not mutually exclusive.

Due to this problem in the training data, the ML model is unable to accurately derive how to predict the correct class.

These cases of class ambiguity in the training data can negatively influence to classificiation ability of the ML model, because if there are two potentially correct classes for a ticket, there is always one class that is less suitable and therefore incorrect.

Table 8.5: Examples of Ambiguous Classes and Misclassification (Category)

| Ticket Text Example | True Label | Predicted Label |
|---|---|---|
| "Need new mobile" | Phone | Phone |
| "Put the mobile in the inventory" | Inventory | Phone |
| "Need a new mouse" | Purchase | Purchase |
| "VPN is not working" | Admin | Admin |
| "Need access rights for Jira" | Admin | Purchase |
| "Need access to CRM" | Onboarding | Admin |

## 5.2   Classification by Escalation Type

The research findings for the classification by escalation type, as presented in Table 8.6, shows that only the balanced dataset led to good performance. M3 with an imbalanced dataset (4,265 tickets) achieved P = 0.623 percent, R = 0.623, and F1 = 0.623 and thus "OK" performance only. For M4 with a balanced dataset (3,443 tickets), the performance significantly increased to P = 0.881, R = 0.881, and F1 = 0.881 and can be qualified as "good". So M4 reached substantially better classification performance (P → +25%, R → +25%, F → +25%), compared to M3, despite using a smaller dataset.

Table 8.6: Escalation ML Model Performance Metrics

| Model | Class distribution | Classification | No. of tickets | P | R | F1 |
|---|---|---|---|---|---|---|
| M3 | Imbalanced | Escalation | 4,265 | 0.623 | 0.623 | 0.623 |
| M4 | Balanced | Escalation | 3,443 | 0.881 | 0.881 | 0.881 |

As shown in Figure 8.7, for M3, 99 percent of tickets to be escalated were incorrectly classified as "Don't escalate". However, 99 percent of the tickets with the true label "Don't escalate" were classified correctly. The effect of imbalanced datasets is relatively clear here, as this example only contained a very small number of escalation cases that could be used to train the model. In contrast, the balanced M4 led to 16 percent misclassification for "Escalate", or confusion with "Don't Escalate", and only 8 percent misclassification for 'Don't escalate'.

| M3 | Predicted Label | | M4 | Predicted Label | |
|---|---|---|---|---|---|
| True Label | Escalate | Don't escalate | True Label | Escalate | Don't escalate |
| Escalate | 1% | 99% | Escalate | 84% | 16% |
| Don't escalate | 1% | 99% | Don't escalate | 8% | 92% |

Figure 8.7: Confusion Matrix Escalation (M1 and M2)

Like the classification by category, the training data was ambiguously classified by a human. The example in Table 8.7 shows, that eventhough both tickets are about installing "software" or an "addon", which is technically the same, two separate classes were used. This leads to the conclusion, that contextual information in the training data is required to derive a logical pattern. Therefore, a ML model derived the wrong pattern and there incorrectly classified this ticket.

Table 8.7: Ambiguous Classes and Misclassification (Escalation)

| Example | True Label | Predicted Label |
|---|---|---|
| "Please install addon " | Don't escalate | Don't escalate |
| "Please install software" | Escalate | Don't escalate |

## 6. Discussion

### 6.1 Applicability for Support Teams

The literature review first showed that little research had been published on the applicability and performance of AutoML methods for IT support ticket classification ($\rightarrow$RQ1). In this paper, we could demonstrate that AutoML (here Google's Vertex AI) can be used for automated ticket classification by ticket category and escalation type classifications, with a good classification performance ($\rightarrow$RQ2). In addition, it could be shown that the performance of the AutoML models strongly depends on the dataset, whereby the amount of training data is less critical than a dataset balanced concerning the classes it contains (→RQ3). The model trained with the balanced dataset (see section 5.1) and 996 tickets for the ticket category classification resulted in the best performance (F1=88%). The performance metric of F1 was even better than the results of previous research with similar amounts of data (e.g., F1=0.86 for a model with 732 tickets by Fuchs et al., 2022a). Compared to the approximated human accuracy of 85 percent (Mandal et al., 2019), this is a performance increase of three percent. However, it must be said that ML models trained with more advanced data preparation steps and larger datasets outperform our results (Goby et al., 2016). Classification by escalation with the balanced ML model M4 (3443 tickets) achieves a classification performance similar to comparable preceding research studies (→RQ2, RQ3). The performance is not directly comparable to research that uses accuracy as an evaluation metric (Jyotheeswar et al., 2020; Yayah et al., 2021). This is because accuracy as a metric is inaccurate for highly imbalanced datasets and prone to interpretation errors. Moreover, our study could not use accuracy to compare performance results as Google's Vertex AI does not output this metric.

## 6.2    Implications for ML Model Development

The final research question was to answer the study's implications for developing ML models for ticket classification (→RQ4). Our research underlines the importance of data quality and shows that the classifications predicted by AutoML models are only as good as the data used for training them. Furthermore, our research results suggest that emphasis should be placed on the classes' quality and uniqueness in the training data to prevent misclassification, as measured in the confusion matrix. Additionally, class balance seems to play a substantial role in classification performance. Further, this research showed that a small training dataset size of about a thousand tickets could be sufficient to achieve good results. However, as other research shows that performance typically increases with the dataset's size, the classification models' performance should increase with more training data. In addition, larger data sets also offer more opportunities for qualitative data selection and cleaning. More advanced data preparation techniques should be tested in future work, e.g., oversampling to improve class balance and topic modeling to derive new classes, and other approaches to improve class balance and uniqueness. Moreover, different AutoML optimization scenarios should be evaluated, e.g., model optimization for high recall instead of F1. Despite the promising results, it must be noted that AutoML, like some other ML and DL models, produces a black box as a classification system. If non-experts generate such models, additional risks may arise. For example, AutoML users might rely too much on the results without being able to challenge or question them adequately. Therefore, extensive tests and validations of the models derived are necessary before they can be applied in practice.

## 6.3    Limitations and Future Research

While this research shows that AutoML is a way to democratize the development of ML models to build an automated ticket classification with good results, there are still some open questions. Therefore, when evaluating the results, it is important to consider the following limitations of our study: Only "Title" and "Description" were used as features, even though more data was available. No oversampling techniques were used to increase class balance and increase examples of minority classes. Only VertexAI was used – no other AutoML solution. The AutoML results were not benchmarked with traditional ML approaches with the same dataset. Therefore, performance comparison to the existing research is only possible to a limited extent. No acceptance and usability tests were conducted for AutoML as a solution to develop a ML classification solution by IT customer service personnel. Finally, the feasibility of integrating the ML model and model updates into existing STS industry solutions was not investigated.

All these areas are potential research fields to be explored by future research. Moreover, it should be mentioned that our results also depend on the dataset used and can, therefore, only be generalized to a limited extent. Again, further research is needed to assess the power of AutoML for the development of automation approaches for classifying IT support tickets in different application domains and business contexts.

## 7.    Conclusion

AI-supported ticket classification in customer support is a developing area with the potential to increase efficiency and reduce workload for support departments. Past research shows that manual classification is error-prone and AI-based classification with ML can improve the classification quality. This research aimed to answer whether AutoML can be used to train an ML model for ticket classification because traditional ML requires highly specialized knowledge and skills in programming and ML. Until now, very little research has been conducted in this research area. This research shows that AutoML can enable non-ML experts to develop ML models with less than 1,000 tickets for ML model training to achieve good prediction quality. The results imply that data quality, selection, and preparation are more important than the dataset size for ML model development. Still, dataset size is a relevant factor to take into consideration. However, further research is needed to specify the data preparation requirements for AutoML further. There is also a need for further research on implementing the obtained models in industrial STS and achieving a reliable classification and acceptance by the customer service staff. Based on this, it is also necessary to investigate how ML models and automated classification can improve customer service processes and sustainably improve CX.